\documentclass[sigconf]{acmart}

\AtBeginDocument{%
  }
    
\AtBeginDocument{%
  }

\setcopyright{acmlicensed}
\copyrightyear{2026}
\acmYear{2026}
\setcopyright{cc}
\setcctype{by-nc-nd}
\acmConference[DAC '26]{63rd ACM/IEEE Design Automation Conference}{July 26--29, 2026}{Long Beach, CA, USA}
\acmBooktitle{63rd ACM/IEEE Design Automation Conference (DAC '26), July 26--29, 2026, Long Beach, CA, USA}
\acmDOI{10.1145/3770743.3804155}
\acmISBN{979-8-4007-2254-7/2026/07}

\usepackage{booktabs}   
\usepackage{algorithm}  
\usepackage{algpseudocode} 
\usepackage{amsmath}    
\usepackage{amsfonts}   

\usepackage{xcolor} 
\usepackage{etoolbox} 
\AtBeginDocument{%
}

\usepackage{graphicx}

\usepackage{graphicx}
\usepackage{hyperref}

\begin{document}

\title{ActionFlow:  A Pipelined Action Acceleration for Vision Language Models on Edge}

\author{
    Yuntao Dai$^{1,2,\dagger}$, 
    Hang Gu$^{1,2,\dagger}$, 
    Teng Wang$^{1,2,*}$, 
    Qianyu Cheng$^{1,2}$, 
    Yifei Zheng$^{1}$,  
    Zhiyong Qiu$^{3}$, \\
    Lei Gong$^{1,2}$,
    Wenqi Lou$^{1,2}$, 
    Xuehai Zhou$^{1,2}$
}


\affiliation{%
  \institution{$^1$University of Science and Technology of China, Hefei, China}
  \country{}
}
\affiliation{%
  \institution{$^2$Suzhou Institute for Advanced Research, University of Science and Technology of China, Suzhou, China}
  \country{}
}
\affiliation{%
  \institution{$^3$IEIT SYSTEMS Co., Ltd., Beijing, China}
  \country{}
}
\affiliation{%
  \institution{ $^\dagger$Both authors contributed equally to this work $^*$Corresponding author}
  \country{}
}

\email{ {frankd35, hanggu, qycheng, yiwaiii}@mail.ustc.edu.cn, {wangt635,leigong0203, louwenqi, xhzhou}@ustc.edu.cn}

\email{qiuzhiyong@ieisystem.com}

\newcommand{\authmark}[1]{\texorpdfstring{$^{#1}$}{}}
\renewcommand{\authors}{Yuntao Dai\authmark{1,2,\dagger}, Hang Gu\authmark{1,2,\dagger}, Teng Wang\authmark{1,2,*}, Qianyu Cheng\authmark{1,2}, Yifei Zheng\authmark{1}, Zhiyong Qiu\authmark{3}, Lei Gong\authmark{1,2}, Wenqi Lou\authmark{1,2}, Xuehai Zhou\authmark{1,2}}

\renewcommand{\shortauthors}{Dai et al.}

\begin{abstract}
Vision-Language-Action (VLA) models have emerged as a unified paradigm for robotic perception and control, enabling emergent generalization and long-horizon task execution. However, their deployment in dynamic, real-world environments is severely hindered by high inference latency.
While smooth robotic interaction requires control frequencies of 20--30 Hz, current VLA models typically operate at only 3--5 Hz on edge devices due to the memory-bound nature of autoregressive decoding. Existing optimizations often require extensive retraining or compromise model accuracy.

To bridge this gap, we introduce \textbf{ActionFlow}, a system-level inference framework tailored for resource-constrained edge platforms.
At the core of ActionFlow is a \textbf{Cross-Request Pipelining} strategy, a novel scheduler that redefines VLA inference as a macro-pipeline of micro-requests.
The strategy intelligently batches memory-bound Decode phases with compute-bound Prefill phases across continuous time steps to maximize hardware utilization.
Furthermore, to support this scheduling, we propose a \textbf{Cross-Request State Packed Forward} operator and a \textbf{Unified KV Ring Buffer}, which fuse fragmented memory operations into efficient dense computations.
Experimental results demonstrate that ActionFlow achieves a \textbf{2.55$\times$} improvement in FPS on the OpenVLA-7B model without retraining, enabling real-time dynamic manipulation on edge hardware.
Our work is available at https://github.com/Frankd35/Action-Flow.
\end{abstract}


\maketitle

\section{Introduction}

Vision-Language-Action (VLA) models represent a paradigm shift in embodied intelligence, unifying visual perception, language instructions, and action outputs within a single, end-to-end sequence modeling framework. By discarding the traditional, modular "sense-plan-act" pipeline, VLAs model the relationship between world states and future actions directly. \cite{sapkota2025vision} This approach has unlocked transformative capabilities. These models exhibit emergent generalization to unseen objects and scenarios, comprehend complex and abstract natural language, and successfully execute long-horizon, multi-step tasks. These breakthroughs are actively driving exploration in applications from home assistant robots to autonomous industrial manipulation \cite{li2024robonurse}.

Despite their conceptual power, the practical deployment of VLAs is severely constrained by high computational latency. It is widely accepted in robotics that control loops involving dynamic interactions—such as obstacle avoidance or human-robot collaboration—require a frequency of at least 10 Hz to ensure stability. Frequencies of 20-30 Hz are considered ideal for fluid and safe interaction \cite{stelmack2024satisfying, ashtiani2021hybrid}. Latency below this threshold makes a robot "sluggish," unable to react to environmental changes (e.g., grasping a moving object or avoiding a sudden obstacle) and can even lead to control-system oscillations. 

\begin{figure}[t]
    \centering
    \includegraphics[width=1\linewidth]{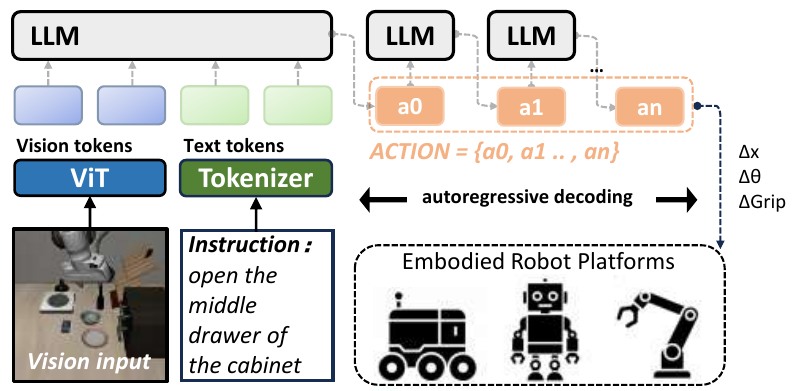}
    \caption{The process of Vision Language Model/Action}
    \label{fig:vla-overview}
\end{figure}

However, mainstream VLAs, which rely on autoregressive inference, as shown in Figure\ref{fig:vla-overview}, typically operate at a mere 3-5 frames per second (FPS) \cite{openvla-oft}. This performance gap is exacerbated on resource-constrained edge devices; for instance, the 7B-parameter OpenVLA model achieves only 3 FPS on a Jetson AGX Orin platform, even with INT4 quantization. This massive performance gap (3 FPS vs. 30 Hz) renders VLAs unsuitable for real-world dynamic tasks, relegating them to slow, turn-based scenarios.

The performance bottleneck originates from the inefficient autoregressive decode phase of the LLM backbone. Existing optimization strategies fail to address this problem in the VLA context. Traditional LLM inference engines are designed to maximize throughput in multi-user, server-side scenarios via request batching; however, they do not address the single-user, low-latency requirements of robotics. Edge-focused optimizations, such as quantization and distillation, reduce model size but do not alter the serial, memory-bound nature of the token-by-token decode loop. Meanwhile, VLA-specific algorithmic optimizations, such as parallel decoding \cite{PD-VLA}, require extensive retraining and risk degrading task performance. Our work fills this critical gap by proposing the first purely systems-level solution that accelerates inference without retraining. Our key insight is that, although a VLA operates as a single external request stream, generating a complete response internally involves a sequence of micro-requests (specifically, one Prefill and multiple Decode operations). This observation suggests that applying batching to this internal sequence may further enhance computational efficiency.

Based on this insight, we propose a novel scheduler using a cross-request pipelining strategy. The scheduler pipelines Prefill and Decode micro-requests between continuous requests, batching multiple sequential, memory-bound Decode micro-requests with a single compute-bound Prefill micro-request into a single compute-bound operation. To enable efficient pipeline execution, we design the Cross-Request State Packed Forward operator to optimize packing computation, which is ultimately integrated into our inference framework, ActionFlow. Applying ActionFlow, our system achieves a 2.55x improvement in FPS on the OpenVLA 7B model.

Our contributions are summarized as follows:
\begin{itemize}
    \item We propose a cross-request pipelining strategy. This strategy treats a single VLA task as a macro-pipeline, batching internal Prefill and Decode micro-requests for efficient computation.
    \item We design and implement a custom-fused "Cross-Request State Packed Forward" operator that aggregates a series of inefficient, memory-bound matrix-vector operations from the Decode phase into a single, compute-bound matrix-matrix operation.
    \item We build ActionFlow, an end-to-end inference framework optimized for resource-constrained edge devices, addressing the gap in systems-level optimization for efficient VLA deployment.
\end{itemize}

\section{Background and Motivation}

\subsection{VLA Architectures}
Vision-Language-Action (VLA) models, such as LLaVA \cite{liu2023llava}, RT-2 \cite{zitkovich2023rt}, and OpenVLA \cite{openvla}, have become central to embodied AI by integrating visual perception with natural language instructions to directly generate executable robot actions. These models are typically built upon pre-trained multimodal large models and fine-tuned on large-scale robotic trajectory datasets to enable end-to-end mapping from observations to actions. Most state-of-the-art VLAs follow an autoregressive generation paradigm—such as RT-2 and OpenVLA—where actions are discretized into tokens and sequentially decoded using a language model. Specifically, at each step, the model generates one action token conditioned on previously emitted tokens, creating a strict sequential dependency. This token-by-token generation process results in inference latency that scales linearly with sequence length, making it incompatible with the high-frequency, low-latency control required for real-time robotic tasks and significantly limiting their practical deployment in dynamic environments.

\subsection{VLA accelerations}
To address this latency challenge, we first examine general LLM optimization techniques, which are predominantly designed for data center scenarios. These techniques have achieved immense success in improving throughput for large-scale services, including memory management optimization \cite{vllm}, scheduling optimization \cite{vllm, orca, sarathi-serve}, IO-Aware Attention \cite{dao2022flashattention, dao2023flashattention2}, various Parallelism strategies \cite{megatron-lm}. 
%
%
Concurrently, a parallel optimization track is designed for resource-constrained edge devices (e.g., mobile phones, robots). Work in this domain has made significant progress in reducing per-inference overhead, primarily including: (1) Quantization \cite{lin2023awq, xiao2023smoothquant}; (2) Speculative Decoding \cite{edge_llm, specexec}; (3) KV Cache Optimization \cite{gqa, liu2024mobilellm}, attention sinks \cite{streamingllm}.

Existing VLA acceleration research has primarily focused on the algorithmic level, aiming to reduce the computational load by modifying the computation task itself. These efforts can be broadly categorized into two types: first, optimizing the existing autoregressive (AR) process; and second, exploring alternatives to autoregression.

In optimizing the AR process, researchers have explored various paths. One path is to reduce data movement and memory access overhead, such as through sparsification \cite{sparsevlm, fastV} or intelligent cache reuse \cite{vla-cache}. Another path is to reduce the number of serial computation steps, using techniques such as speculative decoding \cite{specvla} or dynamic early-exit mechanisms \cite{deervla}. A third path involves model compression and lightweight design \cite{tinyvla, bitvla}.

In exploring alternatives to AR, the research is more diverse. Some work has shifted to non-autoregressive generation paradigms, such as diffusion models \cite{wen2024diffusionvla, li2024cogact} or flow matching \cite{pi_0}. Other efforts have adopted new architectures with linear complexity \cite{robomamba}. Another popular approach is parallel decoding with action chunking \cite{openvla-oft, PD-VLA}. However, these alternatives often introduce new trade-offs: they may require training additional models, increasing system complexity, or cause discontinuous action outputs due to observation lag, affecting control smoothness \cite{pi_0, black2025real}.
%
%
Furthermore, there are explorations into hybrid approaches like fast-in-slow dual systems \cite{fastinslow} or algorithm-architecture co-design \cite{dadu-corki}.

\begin{figure}[t]
    \centering
    \includegraphics[width=0.45\textwidth]{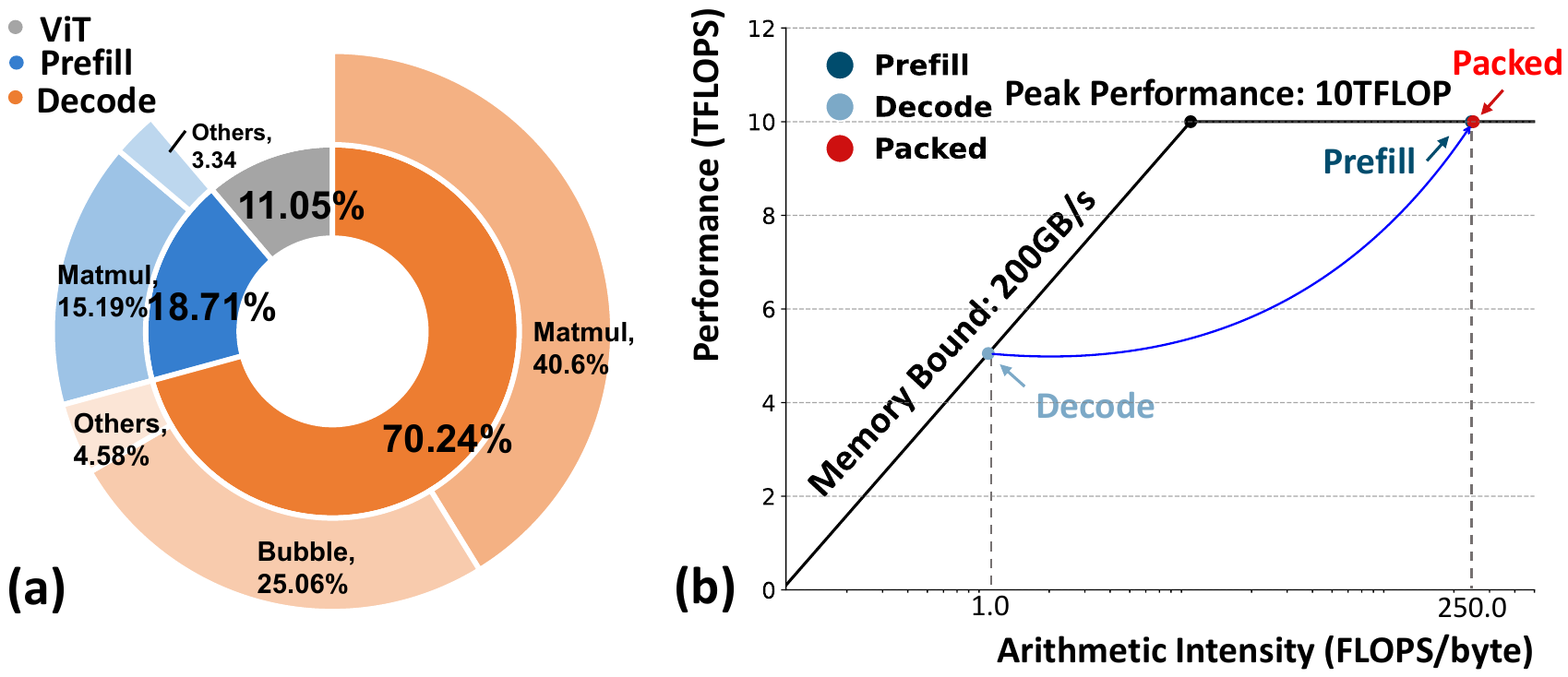}
    \caption{Deconstructing the performance bottleneck. (a) Hierarchical latency breakdown showing VLA inference process (b) Roofline analysis for Jetson AGX Orin platform}
    \label{fig:bottleneck_analysis} 
\end{figure}

\subsection{Motivation}

In summary, we identify a distinct and critical research gap. 1) LLM system-level optimizations (like Continuous Batching) exist, but their application scenario (multi-user vs. single-user) does not match the VLA context. 2) edge-centric optimizations (like quantization), while orthogonal and beneficial, fail to solve the fundamental problem of low hardware utilization during the Decode phase. 3) Existing VLA acceleration efforts remain at the algorithmic level, requiring additional training costs or potentially causing accuracy degradation, and lacking a pure, algorithm-orthogonal system-level acceleration solution.

To pinpoint the true bottleneck, we deconstruct the end-to-end inference latency of the OpenVLA model on its target hardware, as shown in Figure~\ref{fig:bottleneck_analysis}.
%
%
Figure~\ref{fig:bottleneck_analysis}a provides a hierarchical latency breakdown. The inner ring clearly shows that the LLM inference, composed of the Prefill and Decode stages, overwhelmingly dominates the entire pipeline, dwarfing the initial Vision Encoder (ViT) cost. The outer ring further details these stages, revealing that while matrix multiplication (Matmul) is the most time-consuming operation overall, the \textbf{Decode stage} is the single largest source of latency.
%
The Decode stage suffers from two critical, deeply rooted inefficiencies. The first problem is illustrated by the Roofline analysis in Figure~\ref{fig:bottleneck_analysis}b. The key operations in the Decode stage, such as the QKV projection for a single token, possess an extremely low arithmetic intensity (e.g., 1.4096 FLOPs/Byte). This characteristic places them deep within the \textbf{memory-bound} region of the hardware's performance curve, making it impossible to leverage the device's peak compute power (10.0 TFLOPS) and resulting in wasted expensive accelerator resources.
%
%
The second problem is the significant overhead from inefficient kernel execution, as shown visually in the "bubble" portion of Figure~\ref{fig:bottleneck_analysis}a. Because the Decode stage operates token-by-token, it involves launching thousands of small, independent kernels. This serial process incurs substantial scheduling overhead and GPU idle time, further exacerbating the latency bottleneck.
%
%
This analysis allows us to precisely identify the bottleneck's root cause: the inherent nature of autoregressive inference.
%

%
%
Crucially, no existing work addresses the fundamental problem: how to improve the computational efficiency of the Decode stage without external batching requests?

The work presented in this paper directly addresses this gap. We propose the a system-level scheduling solution that accelerates the inference process by creating batching opportunities within a single request. Since the VLA control process must continuously generate these actions, we can leverage this natural continuity. Our core motivation is to overlap the compute-intensive Prefill process of the current [Frame T] with the memory-intensive Decode processes from the recent past [Frame T-1, ...] via batching. This enables a pure, system-level acceleration for VLA, without relying on external multi-user requests or modifying the model's algorithm. 
Our method is fully orthogonal to existing optimizations such as quantization and speculative decoding.

\begin{figure}[t]
    \centering
    \includegraphics[width=1\linewidth]{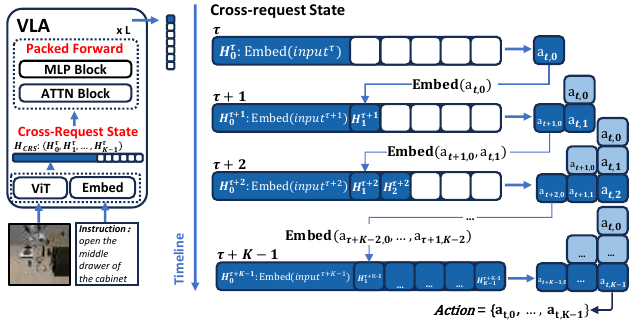}
    \caption{Overview of ActionFlow pipeline}
    \label{fig:actionflow}
\end{figure}

In the specific implementation, we use a variable-length attention operator to efficiently handle the attention computation across different requests, which requires that the KV history data for all K requests be physically stored in a contiguous buffer. The naive method dynamically constructs this contiguous buffer at each layer. This involves gathering data from independent, non-contiguous memory addresses (which store the KV caches of individual requests) and copying it into a newly allocated contiguous memory block. The bottleneck is that such dynamic memory allocation and data copying operations are highly memory bandwidth-intensive. More critically, they require CPU control flow to orchestrate GPU data flow, introducing significant CPU-GPU synchronization bubbles and data movement overheads. To address this issue, we implemented a Cross-Request State Packed Forward method, which restructures the traditional KV cache implementation and uses fused operators to accelerate inference.

\section{ActionFlow Pipeline Design}

\subsection{The Cross-Request Pipelining Strategy}

Vision-Language-Action (VLA) models, such as OpenVLA, form the core of modern embodied agents by mapping visual observations and language instructions to robot actions. These models typically combine a vision encoder with a large language model (LLM) decoder. At each timestep $t$, the agent receives an image $I$ and an instruction $C$. The vision encoder produces visual embeddings $E_{\text{vision}} \in \mathbb{R}^{L_V \times D}$, while the instruction is embedded as $E_{\text{text}} \in \mathbb{R}^{L_T \times D}$. Concatenating them yields the Prefill input $E_t^{\text{prefill}} \in \mathbb{R}^{L_P \times D}$, where $L_V$, $L_T$, and $L_P$ denote the sequence lengths of vision tokens, text tokens, and total prefill tokens, respectively, with $L_P = L_V + L_T$. The LLM then autoregressively generates a fixed-length action token sequence $A_t = \{a_{t,0}, \dots, a_{t,K-1}\}$. We refer to this $K$-step sequence as a \emph{micro-request pipeline}.

%
As illustrated in Figure \ref{fig:actionflow}, we treat the $K$-step generation process (one Prefill, $K-1$ Decodes) as a $K$-stage pipeline. 
%
The core idea is to overlap computation from $K$ consecutive requests in any given computation batch $\tau$. 
%
In each computation batch $\tau$, we pack the Prefill phase of the current request $T_t$ together with the $j$-th Decode phase of $K-1$ historical requests $T_{t-j}$ (for $j=1,\dots,K-1$). This cross-request overlap eliminates pipeline bubbles and maximizes hardware utilization. 
The overall workflow of ActionFlow is formalized in Algorithm \ref{alg:actionflow_algo}.

\begin{algorithm}[t]
\caption{ActionFlow Overall Pipeline}
\label{alg:actionflow_algo}
\begin{algorithmic}[1]
\Require Vision Token $I_t$, Text Token $C_t$ (for request $T_t$), Action Length $K$, Accumulating action sequences for stages $1..K-1$: $\mathcal{A}_{\text{sequences}} \gets [0, A_1, \dots, A_{K-1}]$, Unified KV Ring Buffers: $\mathcal{KV}^{(l)}$ for $l \in [1, N_{\text{layers}}]$ 
\Ensure Generated action sequence $A_{t-(K-1)} \in \mathbb{R}^{K \times 1}$ (for request $T_{t-(K-1)}$)
    \State $\mathcal{A}_{\text{sequences\_next}} \gets []$  

    \State $H \gets \text{Embed}(I_t, C_t, \Call{getLastTokenEach}{\mathcal{A}_{\text{sequences}}})$
    \State $H \gets \Call{PackedForwardEachLayer}{H, \mathcal{KV}}$ 
    \For{$s \gets 0$ to $K-2$} 
        \State $a_{\text{next}} \gets \text{ArgMax}(\text{LMHead}(\text{Norm}(H[s])))$
        \State $\mathcal{A}_{\text{sequences\_next}}.\text{append}(A_{s}.\text{append}(a_{\text{next}}))$
    \EndFor

    \State $a_{\text{final}} \gets \text{ArgMax}(\text{LMHead}(\text{Norm}( H[K-1])))$
    \State $A_{\text{final}} \gets A_{K-1}.\text{append}(a_{\text{final}})$ 

    \State $\mathcal{A}_{\text{sequences}} \gets \mathcal{A}_{\text{sequences\_next}}$ 
    
    \State \Return $A_{\text{final}}$
\end{algorithmic}
\end{algorithm}


The algorithm maintains two persistent states across time steps: 
(1) an accumulation buffer $\mathcal{A}_{\text{sequences}}$, which stores partially generated action sequences for the $K-1$ concurrent requests currently in the pipeline, and 
(2) unified KV cache ring buffers $\mathcal{KV}^{(l)}$ at each transformer layer to manage key-value states across all pipeline stages.

At each time step, we construct a combined input batch by concatenating the current vision and text tokens ($I_t, C_t$) with the latest token retrieved from each active partial sequence in $\mathcal{A}_{\text{sequences}}$. 
This batch undergoes a Packed Execution pass to process all pipeline stages in parallel. 
Subsequently, the generated predictions are appended to their respective partial sequences. 
This causes all sequences to shift to the next stage within the pipeline buffer: the sequence that entered earliest (completed $K$ inference steps) is emitted as the fully generated $A_{\text{final}}$, while the remaining sequences advance for further generation.


\subsection{Cross-Request State and Packed Execution}

To enable this packed execution, we first define the \textbf{Cross-Request State (CRS)}. 
At a computation batch $\tau$, the CRS, denoted $H_{\text{CRS}}^{(\tau)}$, is a single tensor formed by aggregating the $K$ distinct stage inputs $H_s^{(\tau)}$ along the sequence dimension: 
\begin{equation}
H_{\text{CRS}}^{(\tau)} = \text{Aggregate}(H_0^{(\tau)}, H_1^{(\tau)}, \dots, H_{K-1}^{(\tau)})
\end{equation}

Here, $H_0^{(\tau)} = E_t^{\text{prefill}} \in \mathbb{R}^{L_P \times D}$ is the Prefill input for the current timestep $t$. 
The subsequent inputs $H_s^{(\tau)} \in \mathbb{R}^{1 \times D}$ (for $s \in [1, K-1]$) are the token embeddings for the $s$-th decode step of the historical request $T_{t-s}$. 
This aggregation results in the single, packed tensor $H_{\text{CRS}}^{(\tau)}$ with a total sequence length $L_Q = L_P + (K-1)$.

\begin{algorithm}[h] 
\caption{Cross-Request State Packed Forward} 
\label{alg:packed_layer_forward} 
\begin{algorithmic}[1] 
\Require Cross-Request State $H_{\text{CRS}}$, Unified KV Ring Buffer $\mathcal{KV}^{(l)}$, Prefill length $L_0$, sequence lengths $L_s$ for each stage 
\Ensure Output hidden states $H_{\text{CRS}}$ 
\State $H_{\text{norm}} \gets \text{Norm}(H_{\text{CRS}})$ 
\State $Q, K, V \gets \text{QKVProj}(H_{\text{norm}})$
\State \Call{FusedRoPEAndWriteKV}{$Q, K, V, \mathcal{KV}^{(l)}, L_0$}
\State $O_{\text{varlen}} \gets \text{VarlenAttention}(Q_{\text{varlen}}, \mathcal{KV}^{(l)}, L_s)$ 
\State \Call{InPlaceShiftKV}{$\mathcal{KV}^{(l)}, L_s$} 
\State $H_{\text{CRS}} \gets \text{OutputProj}(O_{\text{varlen}}) + H_{\text{CRS}} + \text{MLP}(\text{Norm}(O_{\text{varlen}}))$ 
\State \Return $H_{\text{CRS}}$
\end{algorithmic} 
\end{algorithm}

We feed $H_{\text{CRS}}^{(\tau)}$ as a single monolithic batch through all Transformer layers in one forward pass, as formally described in Algorithm~\ref{alg:packed_layer_forward}. 

%
%
Within each layer, the normalized hidden states are first projected into Queries, Keys, and Values. 
We employ a fused kernel \textsc{FusedRoPEAndWriteKV} to apply Rotary Positional Embeddings and simultaneously deposit the new Key-Value pairs into the \textbf{Unified KV Ring Buffer} $\mathcal{KV}^{(l)}$. 
Self-attention is then computed via a variable-length attention mechanism \textsc{VarlenAttention}, which enables efficient context retrieval from the ring buffer across concurrent pipeline stages with varying history lengths. 
Crucially, before the final projection, an \textsc{InPlaceKVShift} operation updates the ring buffer's sliding window, discarding stale tokens to prepare the memory layout for the next time step.

This design dramatically improves computational efficiency: the $K-1$ tiny $(1 \times D)$ GEMMs from Decode steps—normally memory-bound and ALU-underutilizing—are fused with the medium-sized Prefill GEMM into a single large $(L_Q \times D)$ matrix multiplication. 
The resulting increase in arithmetic intensity (FLOPs/byte) shifts the workload from memory-bound to compute-bound, enabling near-peak GPU utilization. 
Element-wise operations (e.g., LayerNorm, activations) are unaffected by batching and incur no overhead.

However, this batching introduces a key challenge in the self-attention layer: the $K$ segments in $H_{\text{CRS}}^{(\tau)}$ require access to disjoint contextual histories of varying lengths. 
The Prefill segment attends causally over its $L_P$ tokens, while each Decode segment must attend over its own KV cache of logical length $L_P + s$. 
Modern variable-length attention kernels require all KV histories to reside in a physically contiguous memory buffer. 
A naive approach would dynamically gather and copy KV data from $K$ non-contiguous caches before each layer, a process that is not only bandwidth-intensive but also forces CPU-GPU synchronization, erasing performance gains from batching.

\begin{figure}[!t] 
\centering 
\includegraphics[width=1\linewidth]{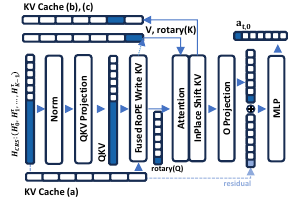} 
\caption{
Workflow of the Packed Layer with Unified KV Ring Buffer. 
The blue blocks represent the KV cache corresponding to the current input frame.
}
\label{fig:operator} 
\end{figure}

\subsection{Efficient KV Cache Management via Kernel Fusion}

To address the memory layout challenge, we introduce a \textbf{Unified KV Ring Buffer}. 
As illustrated in Figure~\ref{fig:operator}, we maintain all active requests' KV states in a single, circular, physically contiguous memory region. 
Although logically partitioned per request, the physical storage satisfies the strict layout requirements of Varlen-Attention kernels. 



Algorithm~\ref{alg:packed_layer_forward} leverages this structure through three coordinated steps, managing the lifecycle of buffer slots as depicted in Figure~\ref{fig:operator}:
First, transitioning from the stale state shown in \textbf{KV Cache (a)}, a fused kernel applies RoPE and writes new keys/values directly to their designated offsets, updating the buffer to the state of \textbf{KV Cache (b)} (Line 3).
Second, a single variable-length attention operation is performed over the entire packed query and the now-contiguous unified buffer (Line 4).
Finally, an in-place shift operation rotates the buffer indices for the next timestep, transitioning the layout to \textbf{KV Cache (c)} (Line 5), thereby eliminating the need for costly memory allocation and CPU intervention.

The first kernel, \textsc{FusedRoPEAndWriteKV}, operates immediately after QKV projection. 
For each Key/Query vector, it applies Rotary Positional Embedding (RoPE) based on its logical position within the original request (e.g., position $L_P + s$ for a token in stage $s$). It directly writes the result to its precomputed physical address in the ring buffer. 
By fusing encoding and write operations, the kernel avoids materializing intermediate RoPE outputs in DRAM, saving significant memory bandwidth and ensuring the Varlen-Attention operator sees a ready-to-use contiguous KV layout.

Between pipeline iterations, the second kernel, \textsc{InPlaceKVShift}, updates the buffer state entirely on-device. 
It performs a \textbf{physical memory copy} to shift the KV data of Stages $0$ through $K-2$ to the memory slots of Stages $1$ through $K-1$, effectively overwriting the evicted oldest request (Stage $K-1$). 
Although involving data movement, this approach offers distinct advantages over standard cache management: it eliminates CPU-GPU synchronization overhead (preventing pipeline bubbles), bypasses the latency of dynamic device memory allocation, and completely avoids memory fragmentation.

Together, these techniques enable ActionFlow to sustain high throughput and low latency in real-time embodied control, transforming the inherently sequential VLA inference into a deeply pipelined, compute-saturated workflow.

\section{Experimental Evaluation}

We conduct a comprehensive experimental evaluation to validate the performance, robustness, and functional correctness of our proposed ActionFlow system.

\subsection{Experimental Setup}

\noindent \textit{Hardware Platforms.} To evaluate the robustness and scalability of our approach across the edge computing spectrum, we conduct experiments on two distinct NVIDIA platforms: 
1) \textbf{NVIDIA Jetson AGX Orin (64GB)}, representing resource-constrained embedded devices typical in mobile robotics; and 
2) \textbf{NVIDIA RTX 5090}, representing high-performance edge workstations with ample memory bandwidth and compute power. 
This dual-platform setup verifies that our optimization is effective across widely varying hardware constraints and architectural generations.


\noindent \textit{Model and Software.} We use the OpenVLA-7B model. Our software stack is built on PyTorch==2.6.0, the Transformers library==4.49.0, and CUDA 12.6.


\noindent \textit{Metrics.} We measure system performance using Frames Per Second (FPS) as the core indicator for the VLA throughput.
To evaluate manipulation capability, we execute multiple tasks across diverse categories (Spatial, Object, Goal, Long) over numerous random seeds. We report the Success Rate to demonstrate that our method incurs no degradation in task accuracy compared to the baseline.

\noindent\textit{Systems for Comparison.}
We compare three implementations:
\begin{itemize}
\item \textbf{Baseline}: OpenVLA-7B Standard autoregressive inference without any system-level optimization. 
\item \textbf{Naive Pipe}: Our cross-request pipelining strategy implemented with dynamic KV cache gathering and CPU-coordinated batching—i.e., without fused operators. 
\item \textbf{ActionFlow}: Full implementation including the pipeline scheduler and our Cross-Request State Packed Forward method with kernel fusion. 
\end{itemize}

\subsection{End-to-End Performance and Ablation Study}

This section evaluates the end-to-end performance of ActionFlow against the baselines to demonstrate its effectiveness and quantify the necessity of our proposed kernel fusions.

Table 1 summarizes the core performance results under a typical workload. The results clearly demonstrate the effectiveness of ActionFlow across both platforms. On the resource-constrained AGX Orin, our system achieves 3.20 FPS, a \textbf{2.56x} speedup over the Baseline (1.25 FPS). On the 5090 data center GPU, we observe a nearly identical \textbf{2.55x} speedup (19.45 FPS vs. 7.62 FPS). This confirms the robustness and cross-architecture validity of our pipelining strategy.

The ablation study, comparing ActionFlow (Ours) to PIPE (Naive), highlights the critical role of our custom fusion operators. The naive implementation, which still suffers from data-copying and synchronization overheads, is significantly outperformed by our full system. On the AGX Orin, ActionFlow provides an additional \textbf{18.5\%} performance uplift over the naive version (3.20 vs. 2.70 FPS). This gap is even more pronounced on the 5090, where our fusion kernels unlock an additional \textbf{24.7\%} performance (19.45 vs. 15.60 FPS). This validates our design, proving that eliminating the synchronization bubbles and data reorganization overheads, as described in Section 3, is essential to realizing the full potential of the PIPE scheduling strategy.

\begin{table}[t]
\caption{End-to-End Performance and Ablation Study}
\resizebox{1.0\linewidth}{!}{
\centering
\begin{tabular}{l|l|c|c|c}
\hline
\textbf{Method} & \textbf{Platform} & \textbf{FPS} & \textbf{Time (ms)} & \textbf{Speedup} \\
\hline
Baseline & AGX Orin & 1.25 $\pm$ 0.01 & 0.8030 $\pm$ 0.0044 & 1.00x \\
Naive Pipe & AGX Orin & 2.70 $\pm$ 0.19 & 0.3731 $\pm$ 0.0299 & 2.16x \\
\textbf{ActionFlow} & \textbf{AGX Orin} & \textbf{3.20 $\pm$ 0.17} & \textbf{0.3131 $\pm$ 0.0170} & \textbf{2.56x} \\
\hline
Baseline & RTX 5090 & 7.62 $\pm$ 0.14 & 0.1312 $\pm$ 0.0024 & 1.00x \\
Naive Pipe & RTX 5090 & 15.60 $\pm$ 0.08 & 0.0641 $\pm$ 0.0003 & 2.04x \\
\textbf{ActionFlow} & \textbf{RTX 5090} & \textbf{19.45 $\pm$ 0.22} & \textbf{0.0514 $\pm$ 0.0006} & \textbf{2.55x} \\
\hline
\end{tabular}
}
\label{tab:ablation}
\end{table}

\begin{figure*}[!htbp]
    \centering
    \includegraphics[width=1\textwidth]{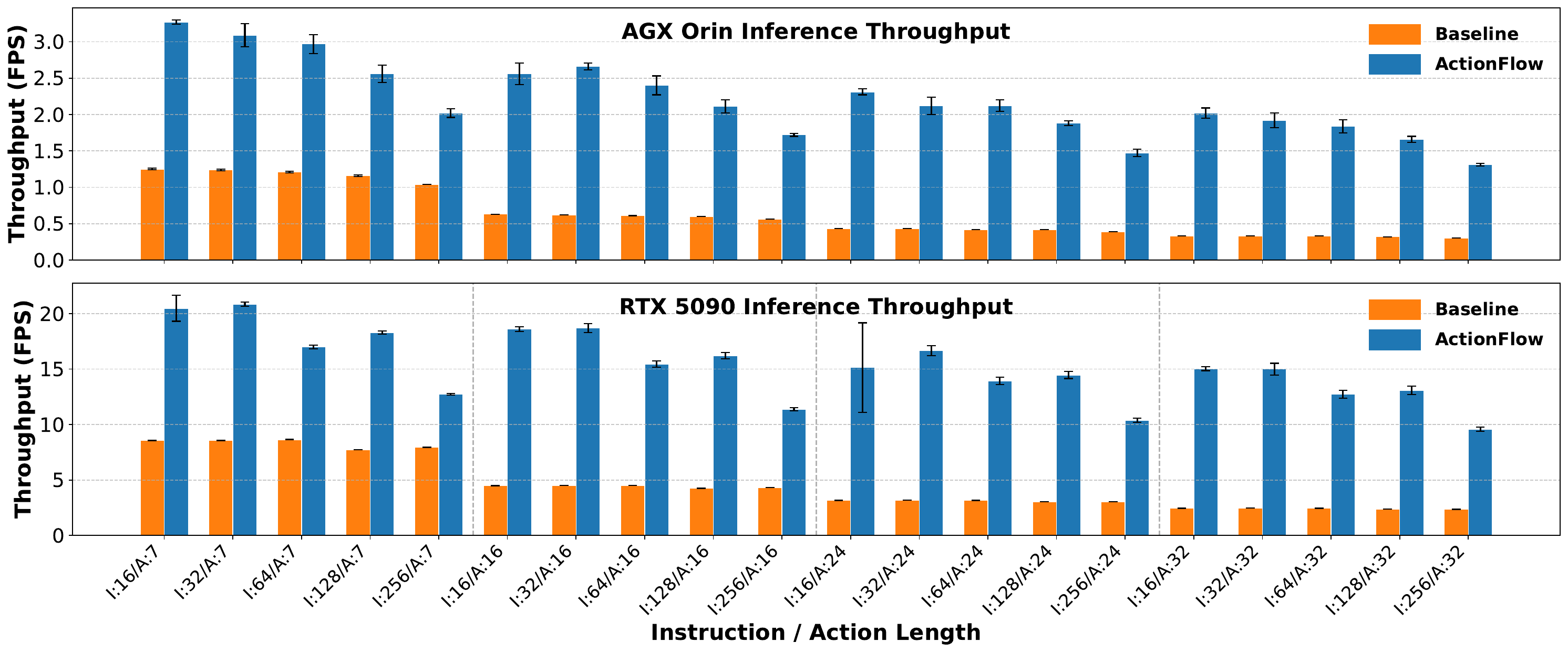}
    \caption{Sensitivity analysis of throughput (FPS) on AGX Orin (top) and RTX 5090 (down) across varying Action Lengths ($K$) and Prefill Lengths ($L_P$).}  
    \label{fig:sensitivity}
\end{figure*}

\subsection{Sensitivity Analysis to Input Workload}


To assess the robustness of ActionFlow, we conduct a sensitivity analysis across a wide range of workloads, varying both the Prefill length ($L_P$) and the Decode length ($K$). Figure \ref{fig:sensitivity} presents a detailed throughput (FPS) comparison between our full system and the Baseline under these varying conditions.

We draw three key conclusions from this analysis. First, ActionFlow performs well across all configurations. As shown in Figure \ref{fig:sensitivity}, our system's throughput (blue bars) is significantly higher than the Baseline's (orange bars) in every action length ($K \in \{7, 16, 24, 32\}$) and for every prefill length(instruction length plus 256 vision tokens) on both hardware platforms.

Second, our \textbf{speedup is most pronounced on heavy workloads}. The advantages of our pipelining strategy are amplified when the task is more demanding. For instance, on the RTX 5090 platform, under the heavy load of $K=32$ and long prefill settings, the Baseline's performance degrades significantly to 2.36 FPS. In contrast, ActionFlow maintains 9.58 FPS, achieving a \textbf{4.06$\times$} speedup. Similarly, on the AGX Orin, we observe a \textbf{4.36$\times$} speedup (1.31 vs. 0.30 FPS) under the heaviest workload.

Third, ActionFlow demonstrates superior \textbf{robustness to the Decode length ($K$)}, which is the primary bottleneck for the Baseline. On the RTX 5090, increasing $K$ from 7 to 32 reduces the Baseline's throughput by approximately 72\% (from 8.64 to 2.45 FPS). Our system, however, only experiences a 25\% drop (from 17.00 to 12.72 FPS). This shows that our cross-request pipelining strategy successfully amortizes the serial overhead of autoregressive generation, making it far more robust for tasks that require more extended action sequences.

\begin{table}[t]
\caption{LIBERO Functional Correctness Validation (Task Success Rate \%)}
\resizebox{1.0\linewidth}{!}{
\centering
\begin{tabular}{l|l|c|c|c|c}
\hline
\textbf{Method} & \textbf{Platform} & \textbf{spatial} & \textbf{object} & \textbf{goal} & \textbf{10} \\
\hline
OpenVLA & RTX 5090 & 84.4\% & 73.8\% & 74.4\% & 51.4\% \\
ActionFlow & RTX 5090 & 84.3\% & 71.2\% & 78.6\% & 53.3\% \\
ActionFlow & AGX Orin & 83.4\% & 68.8\% & 75.6\% & 49.0\% \\
\hline
\end{tabular}
}
\label{tab:correctness}
\end{table}

\subsection{Functional Correctness Validation}

A critical requirement for any system optimization is that it remains functionally lossless. We validate that ActionFlow does not compromise the VLA model's task-completion ability. We use the LIBERO benchmark \cite{liu2023libero} suite to conduct this validation. We execute multiple tasks from different categories (spatial, object, goal, long) across several random seeds, using both the Baseline system and our ActionFlow system on both hardware platforms. We then record and compare the average task success rate.

The analysis of the AGX Orin runs shows that ActionFlow (Ours) maintains success rates equivalent to the Baseline’s functional performance across all task categories (e.g., spatial: 83.4\% vs. 84.4\%; object: 68.8\% vs. 73.8\%; goal: 75.6\% vs. 74.4\%). These minor fluctuations are not considered statistically significant and confirm the functional lossless nature of our system optimization.

\section{Conclusion}

In this paper, we address the critical latency bottleneck hindering the real-world deployment of Vision-Language-Action (VLA) models on resource-constrained edge devices. We introduce \textbf{ActionFlow}, a novel system-level inference framework that fundamentally redefines VLA execution. By deconstructing the autoregressive generation process into a pipeline of micro-requests, our proposed \textbf{Cross-Request Pipelining} strategy effectively overlaps the memory-bound Decode phases of historical steps with the compute-bound Prefill phases of the current step. This architectural shift is enabled by our custom \textbf{Cross-Request State Packed Forward} operator and \textbf{Unified KV Ring Buffer}, which fuse fragmented memory operations into dense computations and eliminate costly synchronization overheads.

Extensive evaluations on the NVIDIA Jetson AGX Orin and RTX 5090 platforms demonstrate that ActionFlow delivers a consistent \textbf{2.55$\times$} speedup for the OpenVLA 7B model, enabling higher frequency control on edge hardware. Crucially, our method incurs no loss in task accuracy and remains orthogonal to algorithmic optimizations such as quantization. By bridging the gap between heavy foundation models and the strict latency requirements of robotic control, ActionFlow paves the way for the deployment of general-purpose embodied agents in dynamic, real-world environments.

\begin{acks}
\small
This work was supported in part by the Strategic Priority Research Program of the Chinese Academy of Sciences, Grant Nos. XDB0660101, XDB0660000, and XDB0660100, in part by the National Natural Science Foundation of China under Grant 62502489, in part by Jiangsu Provincial Natural Science Foundation under Grants BK20241818, BK20251815, and BK20250479, in part by Anhui Provincial Natural Science Foundation under Grant 2508085MF141.
\end{acks}

\bibliographystyle{ACM-Reference-Format}
\bibliography{ref}

\end{document}